\documentclass[conference]{IEEEtran}
\usepackage{times}

\usepackage[numbers]{natbib}
\usepackage{multicol}
\usepackage[bookmarks=true]{hyperref}

\usepackage{mathtools}
\usepackage{amsfonts}  % for \mathbb
\usepackage{graphics}
\usepackage[pdftex]{graphicx}
\usepackage{wrapfig}
\usepackage{color}
\usepackage{algorithmicx}
\usepackage[ruled]{algorithm}
\usepackage{algpseudocode}
\usepackage{amssymb}

\usepackage{wrapfig}

\algnewcommand\algorithmicforeach{\textbf{for each}}
\algdef{S}[FOR]{ForEach}[1]{\algorithmicforeach\ #1\ \algorithmicdo}

\DeclarePairedDelimiterX{\infdivx}[2]{(}{)}{%
  #1\;\delimsize\|\;#2%
}

\usepackage{expl3}
\ExplSyntaxOn
\newcommand\latinabbrev[1]{
  \peek_meaning:NTF . {% Same as \@ifnextchar
    #1\@}%
  { \peek_catcode:NTF a {% Check whether next char has same catcode as \'a, i.e., is a letter
      #1.\@ }%
    {#1.\@}}}
\ExplSyntaxOff
\def\eg{\latinabbrev{e.g}}

\def\etc{\latinabbrev{etc}}
\def\ie{\latinabbrev{i.e}}

\begin{document}

\title{
% Variational Environment Space Analysis: 
Discovering Generalizable Skills via \\ Automated Generation of Diverse Tasks}
% Learning Composable Skills via\\ Automated Generation of Diverse Tasks}

% \title{Task Space Analysis for Diverse Environments: Learning Generalizable Skills by Generating Tasks}
% \title{TSADE: Learning Generalizable Skills by Generating Diverse Tasks}

% Learning Composable Skills via Automated Generation of Diverse Tasks

% You will get a Paper-ID when submitting a pdf file to the conference system
% \author{Author Names Omitted for Anonymous Review. Paper-ID 28}

\author{%
Kuan Fang$^{1}$, %
Yuke Zhu$^{2, 3}$, % 
Silvio Savarese$^{1}$, %
Li Fei-Fei$^{1}$%
\\$^{1}$ Stanford University, %
$^{2}$ UT Austin, %
$^{3}$ Nvidia
}

%\author{\authorblockN{Michael Shell}
%\authorblockA{School of Electrical and\\Computer Engineering\\
%Georgia Institute of Technology\\
%Atlanta, Georgia 30332--0250\\
%Email: mshell@ece.gatech.edu}
%\and
%\authorblockA{Twentieth Century Fox\\
%\and
%\authorblockN{James Kirk\\ and Montgomery Scott}
%\authorblockA{Starfleet Academy\\
%San Francisco, California 96678-2391\\

% avoiding spaces at the end of the author lines is not a problem with
% conference papers because we don't use \thanks or \IEEEmembership

%Homer Simpson\authorrefmark{2},
%Montgomery Scott\authorrefmark{3} and
%Eldon Tyrell\authorrefmark{4}}
%Telephone: (800) 555--1212, Fax: (888) 555--1212}
%\authorblockA{\authorrefmark{4}Tyrell Inc., 123 Replicant Street, Los Angeles, California 90210--4321}}

\maketitle

\begin{abstract}
The learning efficiency and generalization ability of an intelligent agent can be greatly improved by utilizing a useful set of skills. However, the design of robot skills can often be intractable in real-world applications due to the prohibitive amount of effort and expertise that it requires. In this work, we introduce Skill Learning In Diversified Environments (SLIDE), a method to discover generalizable skills via automated generation of a diverse set of tasks. As opposed to prior work on unsupervised discovery of skills which incentivizes the skills to produce different outcomes in the same environment, our method pairs each skill with a unique task produced by a trainable task generator. To encourage generalizable skills to emerge, our method trains each skill to specialize in the paired task and maximizes the diversity of the generated tasks. A task discriminator defined on the robot behaviors in the generated tasks is jointly trained to estimate the evidence lower bound of the diversity objective. The learned skills can then be composed in a hierarchical reinforcement learning algorithm to solve unseen target tasks. We demonstrate that the proposed method can effectively learn a variety of robot skills in two tabletop manipulation domains. Our results suggest that the learned skills can effectively improve the robot's performance in various unseen target tasks compared to existing reinforcement learning and skill learning methods. 

\end{abstract}

\IEEEpeerreviewmaketitle

\section{Introduction}

The ability to acquire diverse and reusable skills is essential for intelligent agents to achieve generalizable autonomy. In well designed tasks, a variety of skills such as grasping~\cite{bohg2014data, kappler2015leveraging, mahler2017dex}, pushing~\cite{Mason1986MechanicsAP, goyal1991planar}, and assembly~\cite{Popov2017DataefficientDR, zakka2020form2fit} can be crafted or learned by a robot. Given a repertoire of skills, previous work has enabled robots to strategically compose these skills to solve novel tasks by performing compositional planning~\cite{kaelbling2011hierarchical, srivastava2014combined,wolfe2010combined, toussaint2018differentiable} and hierarchical reinforcement learning~\cite{barto2003recent, stulp2012reinforcement}. However, the manual design of robot skills often involves a significant amount of time and human expertise. It renders handcrafting the sufficient set of skills intractable for the variability and complexity of real-world tasks.

To reduce the engineering burdens, a plethora of learning-based approaches have aimed to acquire generalizable robot skills from various sources of supervision. Given expert demonstrations or rewards in the target task as the supervision, these skills can be learned through hierarchical reinforcement learning~\cite{Sutton1999BetweenMA, dietterich2000hierarchical} or variational inference~\cite{Pertsch2020AcceleratingRL, fang2019cavin}. Despite their successes, such supervision can be expensive to obtain and the skills trained to specialize in the target tasks might fall short in unseen tasks.

As an alternative, recent works~\cite{Eysenbach2019DiversityIA,Gregor2017VariationalIC, Sharma2020DynamicsAwareUD} have proposed to learn skill-conditioned policies without supervision by incentivizing each skill to choose a sequence of actions that will lead to different outcomes. These methods have opened up a promising direction for unsupervised skill discovery. Nonetheless, most of these approaches gauge the diversity of the skills based on the next states they can reach in the environment, which does not capture the long-term semantics of the robot behaviors. Furthermore, while a set of skills are expected to serve diverse purposes in different scenarios, these methods focus on discovering the skills in a fixed environment. As a result, the learned skills are usually of limited versatility. 

%------------------------------------------------------------------------------%
\begin{figure}[t!]
    \centering
    \includegraphics[width=\linewidth]{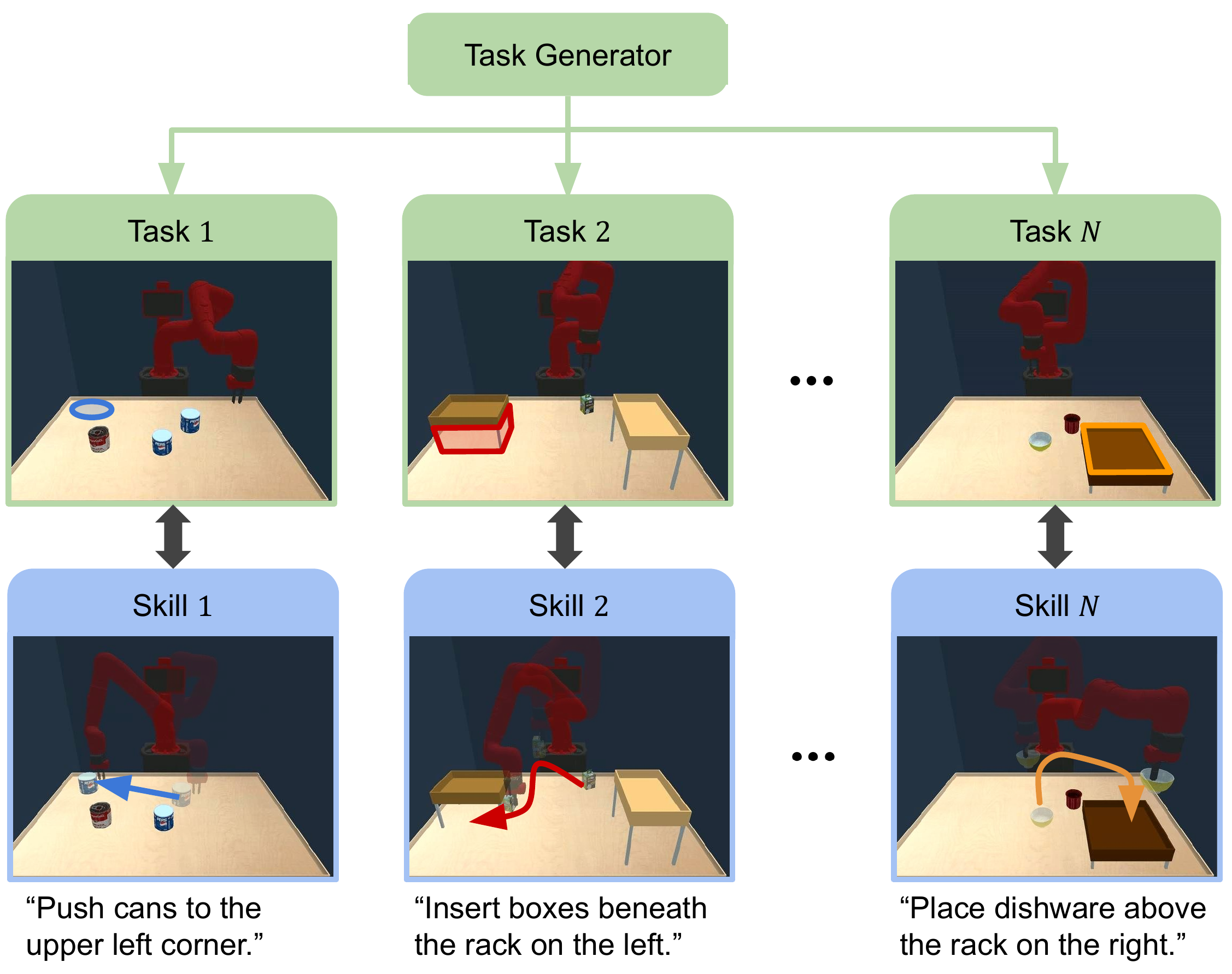}
    \vspace{-6mm}
    \caption{The robot skills are learned through automated generation of tasks in our method. Each skill is paired with a unique task produced by the trainable task generator. The skills are trained to specialize in the paired tasks. We propose to discover a diverse set of skills by diversifying the generated tasks.}
    \label{fig:intro}
    \vspace{-5mm}
\end{figure}

In this paper, we address the problem of learning generalizable skills by automatically generating a variety of tasks. Our key insight is: \emph{a more effective way to acquire a diverse set of skills is to find a diverse set of tasks for training the skills}. In order to create tasks of rich variations, we resort to procedural content generation (PCG), which has been widely used for the automated creation of environments in physics simulations and video games~\cite{Cobbe2019LeveragingPG,risi2019procedural,summerville2018procedural}. Recent works have utilized PCG tools to create benchmarks for robot learning~\cite{summerville2018procedural, risi2019procedural, Cobbe2019LeveragingPG} and automated curricula for challenging tasks~\cite{Portelas2019TeacherAF, fang2020aptgen}. To encourage generalizable skills to emerge, we would like to automate the generation of a diverse set of tasks. 

To this end, we introduce Skill Learning In Diversified Environments (SLIDE), a method that discovers skills by utilizing procedurally generated tasks. In contrast to prior work which learns skills in a fixed environment~\cite{Eysenbach2019DiversityIA,Gregor2017VariationalIC,  Sharma2020DynamicsAwareUD}, our method pairs each skill with a unique task. As shown in Fig.~\ref{fig:intro}, a skill-conditioned \textit{task generator} is trained to create customized tasks for each skill. Instead of directly incentivizing the skills to produce different action distributions or reach distinguishable states, we propose to maximize the diversity of the generated tasks defined by an information-theoretic objective. A \textit{task discriminator} defined on the agent's trajectories is designed to derive evidence lower bound of the objective. By jointly training the task generator, the task discriminator, and the skill policy, our method is able to learn skills of high inter-skill and intra-skill diversities. We further design a hierarchical reinforcement learning (HRL) policy that can efficiently learn to solve unseen target tasks by composing the discovered skills. At each time step, the HRL policy selects the suitable skill to perform various robot behaviors for completing the task goal.

Our experiments are conducted in two tabletop robotic manipulation domains in realistic simulation environments. Our method is able to learn diverse sets of tasks and skills without the notion of target tasks. Although no predefined task semantics are exerted, we found that the learned skills can often be interpreted as semantically meaningful interactions with different types of objects such as pushing, picking, placing, and inserting. Using the learned skills, the hierarchical policy in our method substantially outperforms reinforcement learning and skill learning baselines on unseen target tasks. Videos of generated tasks and learned skills are available at \href{https://kuanfang.github.io/slide/}{\small\texttt{kuanfang.github.io/slide/}}
\section{Related Work}

\subsection{Skill Learning}

There is a large body of work on learning composable robot skills. In hierarchical reinforcement learning (HRL) approaches~\cite{Dayan1992FeudalRL, dietterich2000hierarchical, Sutton2005ReinforcementLA, Kaelbling1996ReinforcementLA, barto2003recent, Thrun1994FindingSI, frans2017meta}, the target tasks can be jointly solved by low-level policies that handle subroutines and high-level policies that learn to select the suitable low-level policies during different stages of the task execution. In these approaches, the skills can be represented by goal-conditioned policies~\cite{Kulkarni2016HierarchicalDR, Deisenroth2014MultitaskPS}, options~\cite{Sutton1999BetweenMA, Precup2000TemporalAI, Bacon2017TheOA}, and dynamic motion primitives~\cite{schaal2006dynamic, Ijspeert2013DynamicalMP} which can be jointly trained with the high-level policies to solve the target task. In addition, pre-defined sub-goals~\cite{heess2017emergence}, hand-designed features~\cite{florensa2017stochastic}, or expert trajectories~\cite{Daniel2013LearningSM, shao2020concept2robot} can be also used for learning the skills. \cite{konidaris2009skill, Bagaria2020OptionDU, Kumar2018ExpandingMS} discovers skills by chaining and scheduling a sequence of policies. Given expert demonstrations, variational inference~\cite{kingma2013auto} can also be applied for learning skill-conditioned and goal-conditioned policies by encoding the trajectories to latent spaces~\cite{Pertsch2020AcceleratingRL, fang2019cavin}. As opposed to these approaches, our method does not utilize any supervision from target tasks or expert demonstrations for learning the skills. Our method resonates with recent works on the unsupervised discovery of diverse behaviors which aim to learn diverse behaviors without the notion of target tasks. \cite{Gregor2017VariationalIC, Eysenbach2019DiversityIA} trains skill-conditioned policies to reach distinguishable states in the environment through mutual information maximization. Similarly, \cite{Sharma2020DynamicsAwareUD} simultaneously discovers predictable behaviors and learns their dynamics. 
\cite{campos2020explore, gangwani2018learning, daniel2012hierarchical, thomas2018disentangling, warde2018unsupervised, lee2019learning} encourage diverse behaviors to emerge by training the agent to maximize the entropy of the action distribution or the coverage of states in the environment. Inspired by the prior work, our method also defines the diversity of skills based on information theory. However, instead of diversifying the robot's behaviors in a fixed environment, our method learns to provide diverse tasks to encourage generalizable skills to emerge.

\subsection{Skill Composition}

Skills can be composed to enable the robot to perform long-term sophisticated interactions with the environment. Conventionally, hierarchical planning~\cite{fikes1971strips, kaelbling2011hierarchical, srivastava2014combined} integrates high-level planning of the sub-task order and the low-level planning of the motions. \cite{mordatch2012discovery, toussaint2018differentiable} enable robots to compose multiple action modes and plan for motion trajectories that can solve sophisticated sequential problems using differentiable physics. Our method does not assume the physical dynamics of the environment to be known beforehand and relies on no predefined action modes. Recently, an increasing number of learning-based methods have been proposed to compose robot skills. In the robotic manipulation domain, several works enable robots to perform tool-use by learning the synergy between grasping and the subsequent manipulation actions~\cite{zeng2018learning, fang2018tog, Zeng2020TossingBotLT} in a fixed order of execution. \cite{bentivegna2001learning, andreas2017modular, denil2017programmable, xu2018neural, danielczuk2019mechanical, huang2019neural, Kurenkov2020VisuomotorMS} composes predefined parameterized motion primitives by exploiting the compositional structure of long-horizon tasks and trains the high-level policy to decide the order and arguments of the primitives. \cite{fernandez2006probabilistic, li2018optimal, Li2019ContextAwarePR, Kurenkov2019ACTeachAB} propose to select or combine a set of sub-optimal policies to effectively learn to solve the target task. As opposed to these methods, our approach does not assume the skills are specified beforehand.

\subsection{Procedural Task Generation}

Procedural content generation (PCG) has been widely used for the automated creation of environments in physics simulations and video games~\citep{summerville2018procedural, risi2019procedural, Cobbe2019LeveragingPG}. Recently, PCG tools have been used to create benchmarks for robot learning and reinforcement learning~\citep{ai2thor, xia2018gibson, savva2019habitat, yu2019meta, James2020RLBenchTR}. Deign of these task environments can be labor intensive and heavily relies on human expertise. Without a carefully designed generation procedure, randomly sampled environments can often be infeasible to solve or trivially easy. Learning-based methods have been recently proposed to generate tasks for training robots and autonomous agents. In \cite{Gravina2019ProceduralCG, Khalifa2020PCGRLPC, Bontrager2020FullyDP}, diverse games and task environments are automatically created for training RL agents. The generated tasks can be used as automated curricula through parameterization of goals~\cite{Forestier2017IntrinsicallyMG, Held2018AutomaticGG, Racanire2020AutomatedCG}, initial states~\cite{wohlke2020performance}, and reward functions~\cite{Gupta2018UnsupervisedMF, Jabri2019UnsupervisedCF}. \cite{OpenAI2019SolvingRC} and \cite{Mehta2019ActiveDR} propose to actively adjust the hyperparameters in physical simulators to alleviate the domain shift by increasingly adding randomization to the physics of the environment.  In \citep{Wang2019POETOC, Wang2020EnhancedPO}, an evolution strategy has been proposed to discover a set of environments by randomly mutating the existing environments in a simulated 2D game. \cite{fang2020aptgen} proposes a general framework that learns to generate new tasks of rich variations with configurable initial state probability, transition probability, and reward function. Our method is similar in spirit to the prior work which aims to learn to procedurally generate tasks for training the robots. However, our method focuses on generating diverse tasks for skill learning. Instead of randomly mutating the tasks or maximizing an indicator of the learning progress of curriculum learning, we train the task generator by defining a diversity objective of the generated tasks. 
\section{Background}
\label{sec:background}

We consider each task as a Markov Decision Process (MDP) denoted by a tuple $M = (\mathcal{S},\mathcal{A}, \rho, P, R, \gamma)$ with state space $\mathcal{S}$, action space $\mathcal{A}$, initial state probability $\rho$, transition probability $P$, reward function $R$, and discount factor $\gamma$. At each time step $t$, the policy $\pi(a | s)$ receives the current state $s_t$ and selects an action $a_t \in \mathcal{A}$ to interact with the environment. A reward $r_t = R(s_t, a_t, s_{t+1})$ is received by the agent at each step as the feedback. To solve the task, the policy is trained to maximize the average cumulative reward $\mathbb{E}[\Sigma_t \gamma^t r_t]$.

Our formulation of skill learning follows the notations from \cite{Eysenbach2019DiversityIA, Sharma2020DynamicsAwareUD}. A latent variable $z \in \mathcal{Z}$ is used to represent the skill index, where $\mathcal{Z}$ can be a discrete or continuous space. A uniform skill prior $p(z)$ is used to sample $z$ during training. Following the practice of \cite{Gregor2017VariationalIC, Eysenbach2019DiversityIA}, we focus on the case of using discrete $z$ due to consideration of learning robustness, while the proposed algorithm can be extended to continuous $z$ with minor modifications. The skill policy $\pi_l(a | s, z; \theta_l)$ is defined to compute the action distribution conditioned on $z$, where $\theta_l$ is the trainable model weight. 

Given the learned skill policy $\pi_l(a | s, z; \theta_l)$, we apply hierarchical reinforcement learning (HRL)~\cite{barto2003recent} to solve the target task $\overline{M}$ by training a high-level policy $\pi_h(z | s; \theta_h)$ with trainable parameters $\theta_h$. The high-level policy is trained to select the suitable skill index $z$ at each time step by maximizing the cumulative reward in the target task. Since we use a discrete $Z$ space in this work, $\pi_\textit{h}$ can be trained with the off-the-shelf Q-learning algorithm~\cite{mnih2015human}. We jointly train the high-level policy and finetune the skill policies for each target task. 

To conduct procedural task generation for skill learning, we consider a task space $\mathcal{T}$ which defines a finite or infinite number of MDPs of similar designs and properties as in \cite{fang2018tog}. We use a multi-dimensional parameter space $\mathcal{W}$ to represent the inter-task variation of $\mathcal{T}$. Given $w \in \mathcal{W}$, a task $M(w)$ can be instantiated in the task space by a predefined mapping $M(\cdot)$. We assume that all tasks share the same $\mathcal{S}$, $\mathcal{A}$ and $\gamma$ such that all policies share the same input and output dimensions. More specifically, each $M(w)$ is defined by a distinct set of $\rho$, $P$, $R$ parameterized by $w$. In general, the target task $\overline{M}$ can be either an instance of unknown parameter $\overline{w} \in \mathcal{W}$ or a task outside of $\mathcal{T}$ but shares the same $\mathcal{S}$ and $\mathcal{A}$. In this work, we consider $\overline{M}$ to be chosen from the parameterized task space.
\section{Skill Learning in Diversified Environments}

We present Skill Learning In Diversified Environments (SLIDE), a method that utilizes procedural task generation for the discovery of generalizable skills. In contrast to prior work which trains the skills in a fixed environment, our method learns to automatically create diversified environments to train the skills. In this section, we first introduce our problem formulation of skill learning through the automated generation of tasks. We then explain our training pipeline and objective functions for learning diverse tasks and skills. 

%------------------------------------------------------------------------------%
\begin{figure}[t!]
    \centering
    \includegraphics[width=0.9\linewidth]{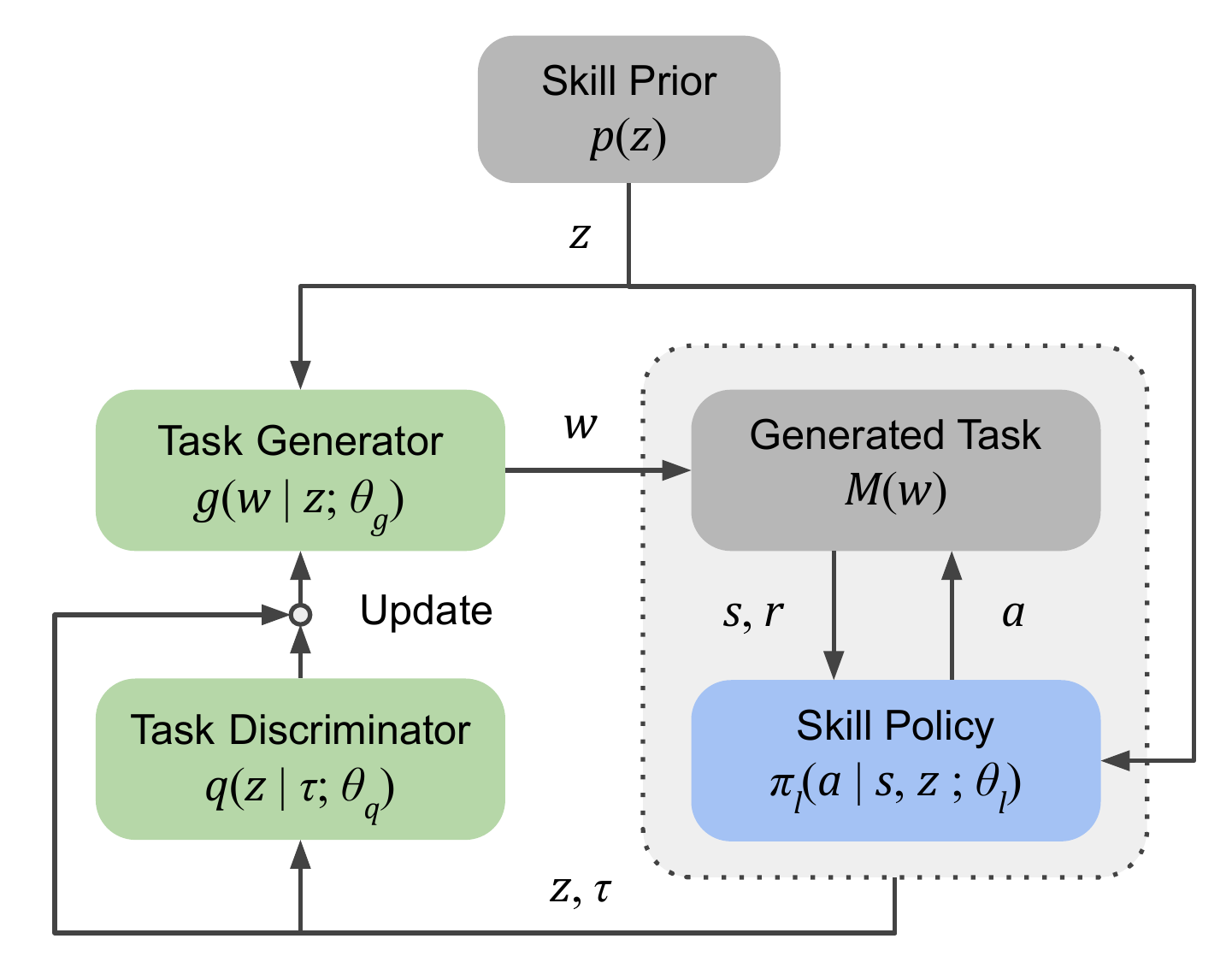}
    \vspace{-3mm}
    \caption{Skill Learning In Diversified Environments (SLIDE). Conditioned on the sampled skill index $z$, the skill policy is trained in tasks procedurally created by the task generator. The task generator is trained to maximize an objective defined by the collected trajectory $\tau$ and the task discriminator. Task generators and discriminators are marked in green, the policy is marked in blue, and predefined modules are marked in grey. }
    \label{fig:method-slide}
    \vspace{-3mm}
\end{figure}

%------------------------------------------------------------------------------%
\subsection{Procedural Task Generation for Skill Learning}

Our method integrates skill discovery and procedural task generation. Instead of learning all skills in the same environment, we pair each skill with a task for it to specialize. Our key insight is that a diverse set of skills will emerge by learning in a diverse set of tasks where each task requires unique robot behaviors to solve. 

Given a parameterized task space $M(\cdot)$, a task can be represented by a task parameter $w$ sampled from the task parameter space $\mathcal{W}$ as $M(w)$. However, each $w$ often defines a specific configuration of the environment and the task goal. Training each skill for a single $M(w)$ can lead to limited generalization capability and a prohibitive amount of task-skill pairs need to be created to cover the useful robot behaviors in the task space. Alternatively, we can partition $\mathcal{W}$ into multiple modes and pair each skill with a mode of tasks. While a hard partition of the task space could be sub-optimal, we would like to extend the notion of using a task distribution $p(w | z)$ conditioned on the skill index $z$. Given the same $z$, $p(w | z)$ is supposed to sample $w$'s that will lead to the same task semantics and require similar robot behaviors to solve. 

To learn to generate suitable tasks for discovery of skills, we define a skill-conditioned task generator $g(w | z; \theta_g)$ to capture the task distribution $p(w | z)$, where $\theta_g$ are learnable model parameters. As opposed to prior work on curriculum learning and automated task generation~\cite{Held2018AutomaticGG, matiisen2019teacher, Portelas2019TeacherAF, Racanire2020AutomatedCG, fang2020aptgen} which create a sequence of goals or tasks as curricula, the task generator in our method aims to learn a set of task distributions for learning a diverse set of skills. The input $z$ to our task generator is defined as the skill index which indicates the unique task semantics that the corresponding skill learns to solve, instead of being a random noise used in deep generative models. The task generator $g$ can be defined for an arbitrarily complex task parameter space $\mathcal{W}$ and learns to model a categorical or Gaussian distribution for each discrete and continuous variable respectively. In this work, we implement $g$ as a neural network that takes $z$ as input and predicts the arguments (\ie~logits, means, standard deviations) of the probability distribution for each task parameter.

As shown in Fig.~\ref{fig:method-slide}, the interplay between the task generator $g$ and the skill policy $\pi_l$ can be formulated by a teacher-student paradigm~\cite{matiisen2019teacher}, in which the teacher $g$ learns to provide suitable tasks for the student $\pi_l$ to solve. During training, the policy learns to maximize the cumulative reward in the updated tasks created by the task generator, while the task generator adapts the task distributions based on the robot behaviors resulted from the current policy. The model parameters $\theta_g$ and $\theta_l$ are jointly optimized in this process.

%------------------------------------------------------------------------------%
\begin{algorithm}[t]
\caption{Skill Learning In Diversified Environments}
\begin{algorithmic}[1]
\Require parameterized task space $M(\cdot)$ and skill prior $p(z)$

\State Initialize model parameters $\theta_g$, $\theta_q$, $\theta_l$, $\alpha$

\While{not converged}
    \State Sample skill $z \sim p(z)$
    \State Sample task parameter $w \sim g(w | z; \theta_g)$
    \State Create task $M(w)$ and initialize $s_1 \sim \rho(s_1 | w)$
    \State Start a new trajectory $\tau = \varnothing$
    
    \For{$t = 1, ..., T$}
        \State Sample action $a_t \sim \pi(a_t | s_t, z; \theta_l)$
        \State Step the environment $s_{t+1} \sim P(s_{t+1} | s_t, a_t, w)$
        \State Receive the reward $r_{t} = R(s_t, a_t, s_{t+1}, w)$
        \State Append the trajectory $\tau = \tau \cup \{(s_t, a_t, r_t, s_{t+1})\}$
        \State Update $\theta_l$ to maximize the cumulative reward
    \EndFor
    
    \State Compute the inter-skill diversity using $\log q(z | \tau; \theta_q)$
    \State Update $\theta_q$ to minimize the classification error
    \State Update $\theta_g$ to maximize the objective in Eq.~\ref{eqn:objective_kkt}
    \State Update $\alpha$ by using Eq.~\ref{eqn:objective_alpha}
    
\EndWhile

\end{algorithmic}
\label{algo:main_algorithm}
\end{algorithm}
\setlength{\textfloatsep}{10pt}

%------------------------------------------------------------------------------%
\subsection{Learning to Generate Diverse Tasks}

%------------------------------------------------------------------------------%
\begin{figure}[t!]
    \centering
    \includegraphics[width=0.95\linewidth]{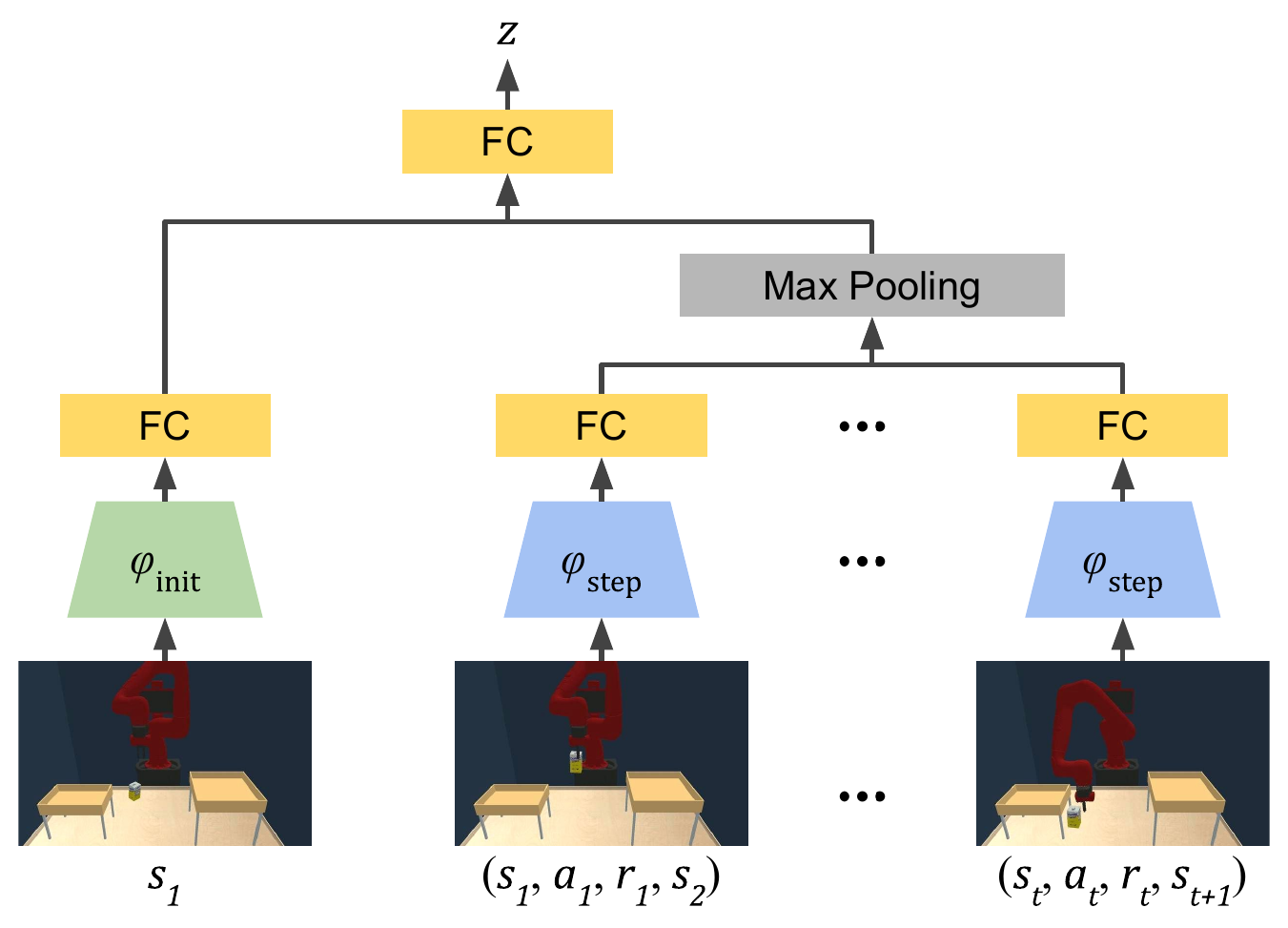}
    \vspace{-3mm}
    \caption{Task discriminator network architecture. The network takes the trajectory collected by each skill as input. It encodes the initial state and each transition and predicts the skill index $z$ from the pooled feature.}
    \vspace{-2mm}
    \label{fig:task_discriminator_network}
\end{figure}

% Desired properties of skills. Inter-task, intra-task variations, usefulness.  
The key to the skill discovery in our method is to design an objective function for training the skill-conditioned tasks generator $g$ to create diverse tasks. To enable generalizable skills to emerge, we argue that both \textit{inter-skill diversity} and \textit{intra-skill diversity} need to be taken into considerations and properly balanced when training the task generator $g$. Inter-skill diversity encourages each task to present unique challenges for the paired skill to solve. While intra-skill diversity gauges the variations of environments that each task can provide. For robotic manipulation tasks, we would like each robot skill to specialize in a different type of interaction (\eg~pushing, grasping, placing, \etc) with a specific type of object. Meanwhile, we expect each skill to have sufficient generalizability to handle the scene variability and task initialization. Lastly, we take the \textit{feasibility} of the tasks into account to prevent learning skills in tasks that are infeasible to solve. 

Inspired by prior work~\cite{Eysenbach2019DiversityIA, Gregor2017VariationalIC}, we define information-theoretic objectives to represent inter-skill and intra-skill diversity. However, instead of directly defining the diversities for the policy, our method aims to diversify the tasks created by the skill-conditioned task generator and the robot behaviors performed by the corresponding skill policy. To this end, we collect the robot's trajectory $\tau = \{ (s_t, a_t, r_t, s_{t+1}) \}_{t=1}^T$ of $T$ steps by sampling different skill index $z$ and define the inter-skill diversity as the mutual information $I(\tau ; z)$ between $\tau$ and $z$. The trajectory $\tau$ is jointly determined by $g$ and $\pi_l$ and provides information about the initialization, dynamics, and rewards of the task. When each skill is trained to specialize in its paired task, the $\tau$ unrolled by each skill captures the semantics of the task. By measuring how well we can infer about $z$ given $\tau$, this term represents the difference among robot behaviors resulted from choosing different $z$. To measure the intra-diversity, we use the conditional entropy $\mathcal{H}(w | z)$ of the task generator $g$. That is, we would like the task parameters associated with each $z$ to be as diverse as possible. To measure the feasibility of task, we compute the average cumulative reward $\mathbb{E}_{p(\tau | z)}[ \Sigma_{t} \gamma^t r_t ]$ that the robot achieves by choosing the skill $z$, where $\mathbb{E}_{p(\tau | z)}[\cdot]$ is a shorthand notation that represents the expectation over distribution of $\tau$ conditioned on $z$. In summary, we train the task generator to maximize: 

\begin{equation}
    J = I(\tau ; z) + \mathcal{H}(w | z) + \mathbb{E}_{p(\tau | z)}[ \Sigma_{t} \gamma^t r_t ].
    \label{eqn:objective_original}
\end{equation}

Since the true posterior $p(z | \tau)$ is unavailable, $I(\tau ; z)$ cannot be evaluated directly. Instead, we introduce a task discriminator $q(z | \tau; \theta_q)$ as an approximation to $p(z | \tau)$. Note that in contrast with the approximate posterior used in \cite{Eysenbach2019DiversityIA} which only takes the next state as the input, $q$ estimates the posterior distribution of $z$ using the information of the full trajectory $\tau$ composed of the sequence of states, actions, and rewards. As shown in Fig.~\ref{fig:task_discriminator_network}, the task discriminator first encodes the initial state and each time step using encoder networks $\phi_\text{init}$ and $\phi_\text{ste}$, then merges the information across time steps to estimate the posterior of $z$. The task discriminator is trained to minimize the classification error given the ground truth $z$ sampled during data collection. As a result, this allows us to solve the above problem by maximizing the variational lower bound of Eq.~\ref{eqn:objective_original}:
\begin{equation}
    J = \mathbb{E}_{p(\tau | z)}[ \log q(z | \tau) ] + \mathcal{H}(w | z) + \mathbb{E}_{p(\tau | z)}[ \Sigma_{t} \gamma^t r_t ].
    \label{eqn:objective_elbo}
\end{equation}
The pseudocode of the SLIDE algorithm is summarized in Algorithm~\ref{algo:main_algorithm}. During training, new tasks are continuously created and used for training the skills. In each new episode, a skill index $z$ is sampled from $p(z)$ for the task generator $g$ to create a new task instance $M(w)$. Then a trajectory $\tau$ is unrolled by the skill policy $\pi_l$. The training alternates among updates of the skill policy, the task discriminator, and the task generator using the collected trajectories. 

%------------------------------------------------------------------------------%
\subsection{Automating Diversity Adjustment}

% Implementation. Adaptive weight. How to balance. Add the term later on. 
The inter-skill diversity and the intra-skill diversity need to be properly balanced to obtain generalizable skills. Hand-tuning a weight in the objective function or simply forcing the diversity to a fixed value would lead to poor solutions since the task generator should be free to explore the parameterized task space before the objective converges. Inspired by \cite{Haarnoja2018SoftAO}, we define $\bar{\mathcal{H}}$ as the target intra-skill diversity for each skill. We use the weight $\alpha$ in the objective to balance the terms. Specifically, we rewrite Eq.~\ref{eqn:objective_elbo} to be:
\begin{equation}
    J = \mathbb{E}_{p(\tau | z)}[ \log q(z | \tau) ] + \alpha \mathcal{H}(w | z) + \mathbb{E}_{p(\tau | z)}[ \Sigma_{t} \gamma^t r_t ].
    \label{eqn:objective_kkt}
\end{equation}
The weight $\alpha$ is constantly updated based on the difference between the evaluated intra-skill diversity $\mathcal{H}(w | z)$ and the chosen $\bar{\mathcal{H}}$. The optimal $\alpha^*$ can be solved as:
\begin{equation}
    \alpha^* = \arg\min_{\alpha} \mathbb {E}_{z \sim p(z)} [ \alpha \mathcal{H}(w | z) - \alpha \bar{\mathcal{H}}].
    \label{eqn:objective_alpha}
\end{equation}
The suitable $\bar{\mathcal{H}}$ depends on the task domain and is chosen to be 3 in our experiments. 
\section{Experiments}

The goal of our experimental evaluation is to answer the following questions: 1) Can SLIDE discover diverse skills by learning to procedurally generate tasks? 2) Can the skills learned by SLIDE be utilized and generalized for learning to solve unseen tasks? 3) How do the design options in SLIDE affect the learned skills and the task performance?

%------------------------------------------------------------------------------%
\subsection{Experiment Setup}

%------------------------------------------------------------------------------%
\begin{figure}[t!]
    \centering
    \includegraphics[width=\linewidth]{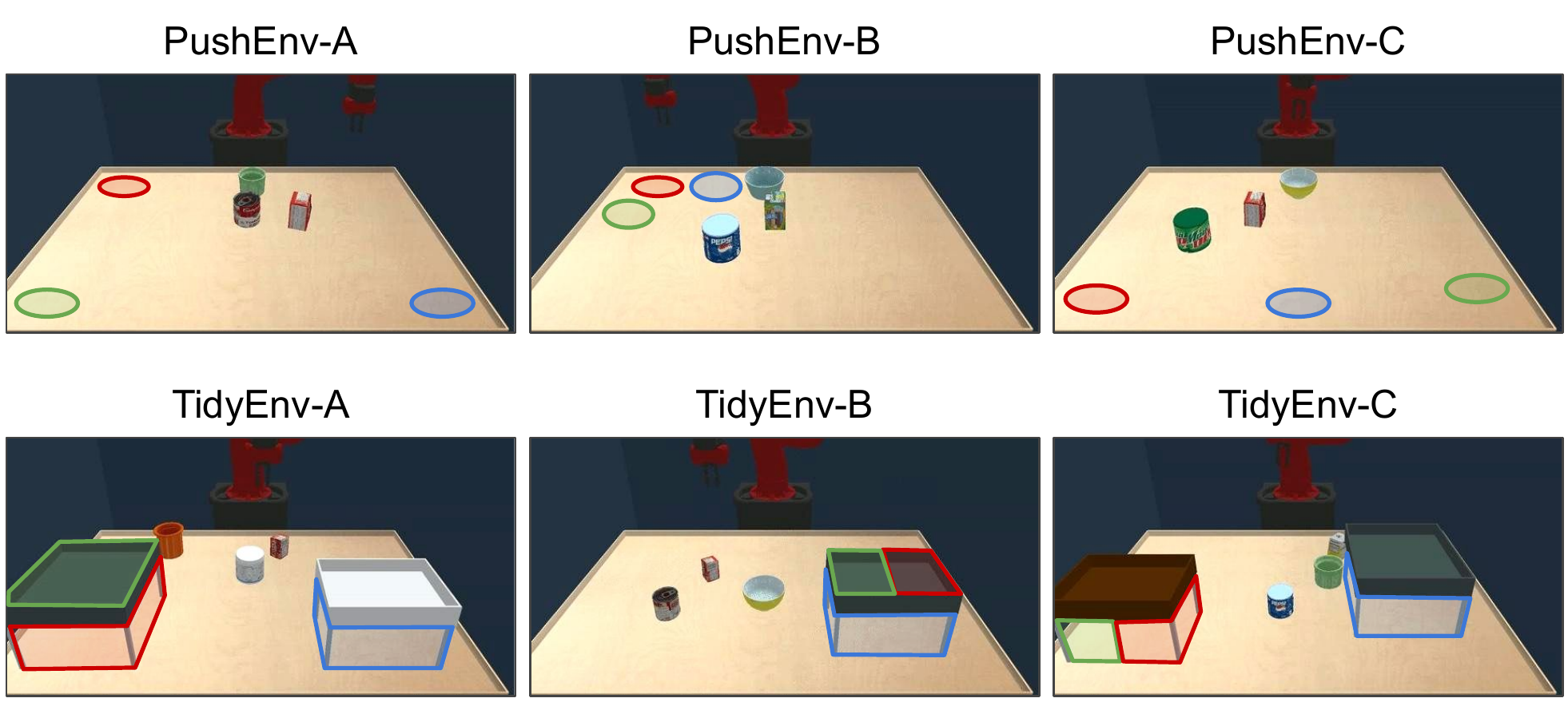}
    \vspace{-5mm}
    \caption{Target tasks in the two domains. Goals of different object categories are indicated as: cans (red), boxes (green), and dishware (blue).}
    % \vspace{-3mm}
    \label{fig:experiments-target-tasks}
\end{figure}

%------------------------------------------------------------------------------%
\begin{figure*}[t!]
    \centering
    \includegraphics[width=\linewidth]{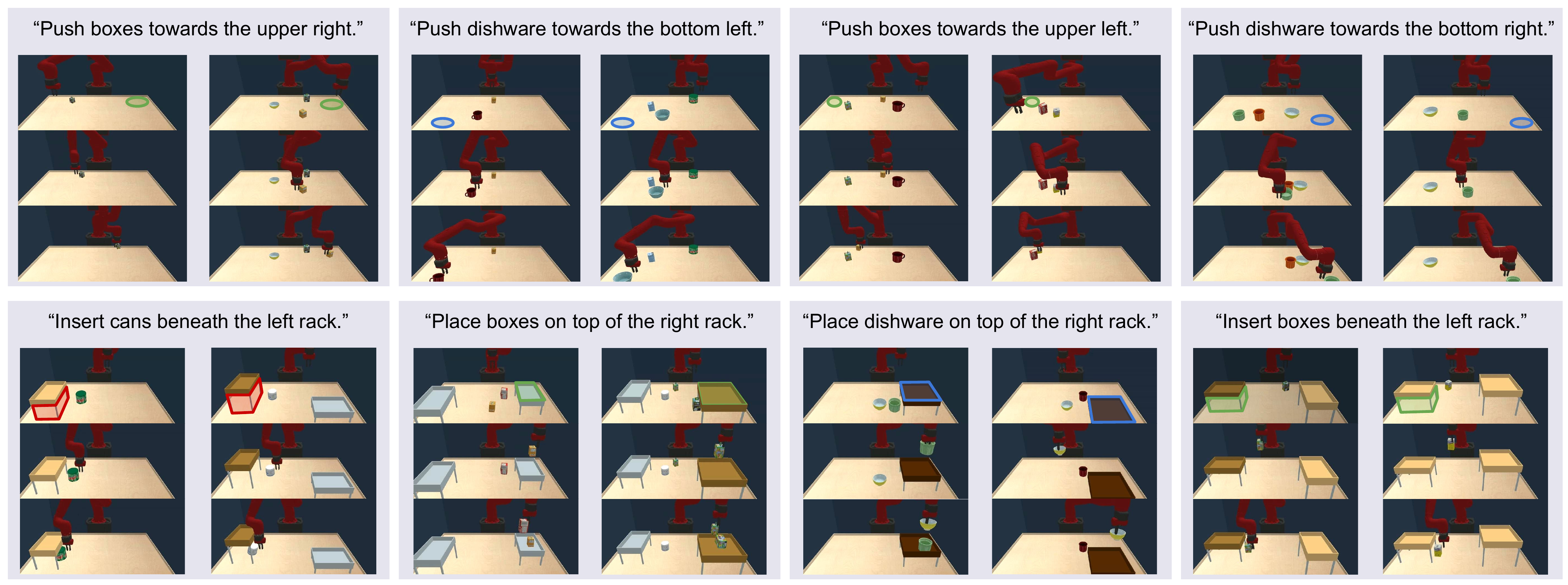}
    \vspace{-5mm}
    \caption{Example tasks and skills discovered by SLIDE. We demonstrate the inter-skill and intra-skill diversity by showing two sampled trajectories associated with the same skill index in each grey block. Each column shows the initialization of the generated task (the top tile) and the execution of the skill (the second and the third tiles). Different colors indicate destinations of different object categories including cans (red), boxes (green), and dishware (blue).}
    \vspace{-3mm}
    \label{fig:experiments-tasks-and-skills}
\end{figure*}

To learn robot skills and evaluate their generalization capability to unseen target tasks, we design two tabletop manipulation domains. Each domain defines a task space that contains various tasks that share the same state and action spaces but different designs of environments and reward functions. The two task spaces are parameterized by multiple discrete and continuous variables to define the initialization, dynamics, and reward functions. We first train our method to discover skills by procedurally generating tasks from the parameterized task spaces without the notion of the target task. Then we train the hierarchical policy to solve each unseen target task by utilizing the skills learned from the same domain.  

As shown in Fig.~\ref{fig:experiments-target-tasks}, our tasks involve a 7-DoF Sawyer robot arm interacting with multiple objects in a configurable table-top environment. The two task domains, which we name \textit{PushEnv} and \textit{TidyEnv}, are adapted from the task design from prior work on tabletop manipulation and object rearrangement~\cite{finn2017deep, fang2019cavin, zhu2020hierarchical, Batra2020RearrangementAC}. The task is simulated by a real-time physics engine~\cite{coumans2019} using the geometry and physical parameters that match a real-world Sawyer robot. In each episode, the environment is initialized with 1-3 movable objects randomly placed on the table. The objects are chosen from 3 categories including cans, boxes, and dishware. The total number, categories, initial placements, and attributes of objects can vary across episodes, so the robot needs to manipulate the objects accordingly. The task requires the robot to move the objects to initially unknown destinations. To complete the task, the robot is supposed to infer the task goals by interacting with the environment and maximize the reward by strategically rearranging the objects. The robot arm operates in a continuous action space through position control, which enables the robot arm to perform a variety of interactions with the objects in the tabletop environments including picking, placing, pushing, \etc. Each episode terminates after 20 steps or when robots collide with the table. More specifically, we describe the parameterization, the state space, and the rewards of the two task domains below.

% (category + scale + mass + friction) * num_bodies + (x, y) * num_body_types
% = 4 * 3 + 2 * 3 = 18
\textbf{PushEnv.} The robot aims to push objects of different categories towards the corresponding locations. In this task domain, the robot can move the end-effector arbitrarily in the constrained 3D space above the table while its fingers are fixed such that the objects cannot be directly picked and placed to the goals. To effectively complete the task, the robot is supposed to alternate between sliding over the table to push target objects towards the goals and lifting the arm to avoid undesired collisions with the distractor objects. The task is parameterized by a totally 15 independent variables which include the object category, object size, goal size, and 2D goal location corresponding to each object category. The robot observes the gripper position as well as the object positions and physical properties but does not know the goal location. The action is defined as a 3-dimensional vector representing the target position that the end-effector is going to be moved to. The reward is defined as the displacement of the object towards the goal and can be either positive or negative.

% (category + scale + mass + friction) * num_bodies + 1 * num_body_types + (w, h, z) * num_racks
% = 4 * 3 + 1 * 3 + 3 * 2 = 21
\textbf{TidyEnv.} The robot is asked to tidy up the table by rearranging objects to specific destinations above or beneath the racks. In addition to moving the end-effector, the robot is enabled to open or close its fingers to pick or place objects. Up to two static racks are attached to the table. Each rack can have varying shapes, heights, and locations across episodes. The objects of each category may have a destination as the upper or lower level of one of the racks or remain on the table surface. To move the objects to the destinations, the robot needs to strategically pick, place, and insert the objects. The task is parameterized by 12 independent variables. Aside from the aforementioned object properties, the parameter includes categorical variables presenting the destination of each object category as well as the height of the rack. The robot observes the gripper position, the finger status, and the object properties. The action is defined as a 4-dimensional vector representing the target end-effector position as well as an additional value chosen between 0 and 1 to indicate whether the robot fingers are closed. The robot receives a sparse reward of 1 when an object is moved to the correct destination. 

\begin{figure*}[t!]
    \centering
    \includegraphics[width=.99\linewidth]{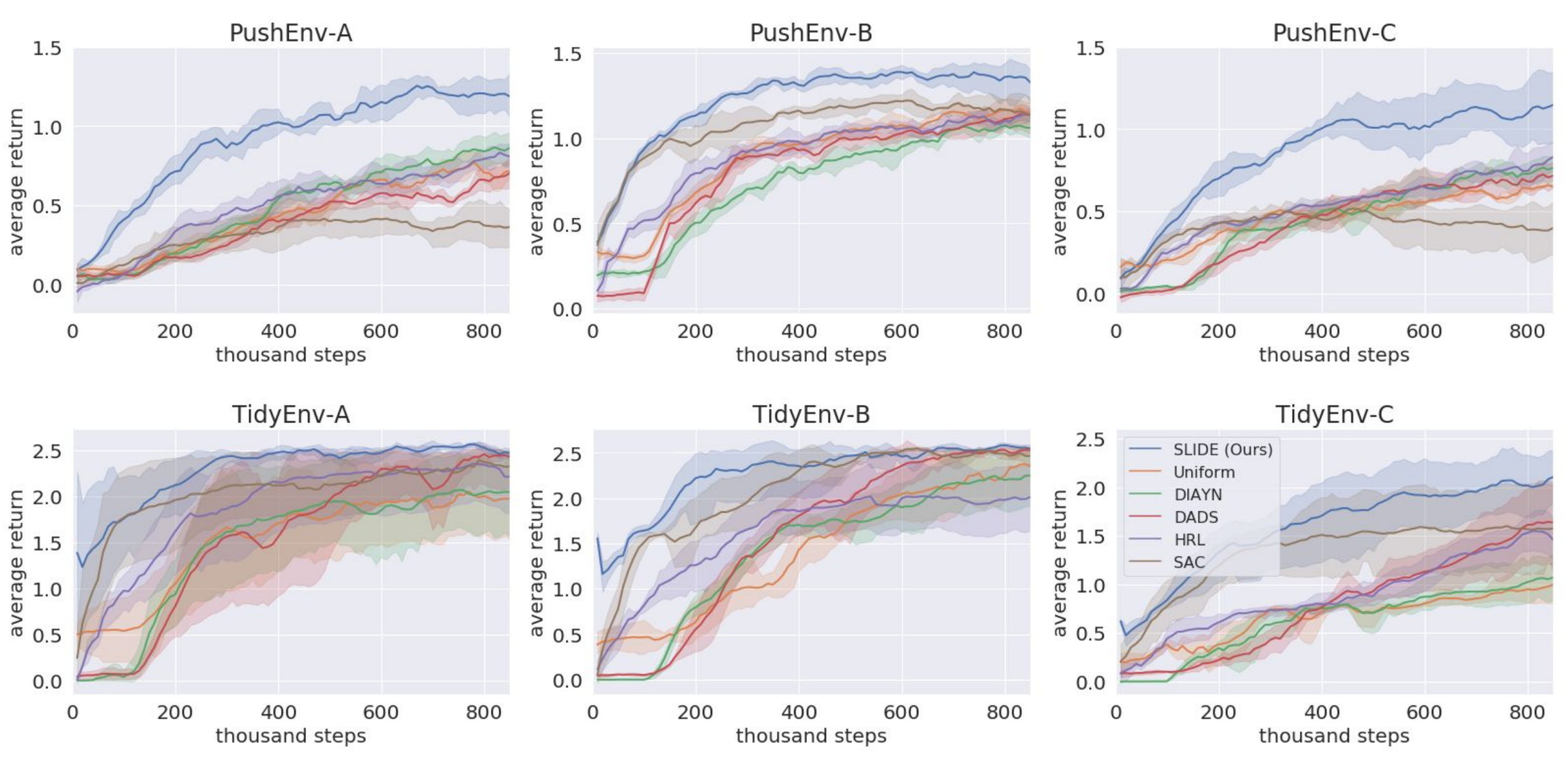}
    \vspace{-3mm}
    \caption{Quantitative results in the target tasks. Given the skills learned by SLIDE, we train the hierarchical policy to solve unseen target tasks. Our method outperforms multiple reinforcement learning and skill learning baselines in terms of the average return. 
    % \yuke{minor: ylim [0, 2.75] and xlim [0, 800], the figures do not seem to be vectorized. Same as Fig 7}
    }
    \vspace{-3mm}
    \label{fig:experiments-comparison}
\end{figure*}

\textbf{Implementation Details.} Soft-Actor-Critic (SAC)~\cite{Haarnoja2018SoftAO} is applied to train the skill policies in the generated tasks. The actor networks and critic networks for SAC and the Q-network for deep Q-learning are implemented with an object-centric feature extractor adapted from \cite{fang2019cavin}. The task discriminator network is adapted from \cite{fang2020aptgen} which first separately encodes each modality of the trajectory and then concatenates the encoded features for each time step in order to eventually predict the skill index $z$ for the input trajectory $\tau$. 64-dimensional fully-connected (FC) layers are used in the network architectures and 64 skills are learned in SLIDE. For all experiments, we use the ADAM optimizer~\citep{kingma2014adam} with learning rate of $3 \times 10^{-4}$, $\beta_1 = 0.9$, $\beta_2 = 0.999$ and the batch size of 128. Hyperparameters are chosen by random search. 

% For SLIDE, we choose the number of skills to be 64 and the target intra-skill diversity to be 3 as the default hyperparameters.

%------------------------------------------------------------------------------%
\subsection{Analysis of Generated Tasks and Learned Skills}

We train SLIDE for 500K iterations and visualize the discovered tasks and skills in Fig.~\ref{fig:experiments-tasks-and-skills}. Each grey block shows two example trajectories associated with a different skill index $z$ in one of the two task domains. The initialization of the generated task (the top tile) and the execution of the skill policy (the second and the third tiles) are demonstrated in each column. Marks of different colors are used to indicate destinations for different object categories including cans (red), boxes (green), and dishware (blue). As shown in the figure, the robot learns to perform a variety of behaviors. Although the task semantics are not predefined in our model, we found that the learned skills can often be interpreted as semantically meaningful interactions with different types of objects such as pushing, picking, placing, and inserting. The task generator usually learns to generate a concentrated distribution of object types and goal locations for the skill to focus on. Given generated tasks of diverging semantics, we found that a single skill suffers to make consistent progress and the task discriminator can hardly identify the skill index based on the resultant trajectories. As a result, the inter-task diversity and the feasibility term often make the distribution of task parameters that are critical to the task semantic quickly converge to a single mode. Meanwhile, the intra-skill diversity encourages the task parameters that have a smaller effect on the robot behaviors to diverge as much as possible. The task parameters often need to co-adapt to create meaningful tasks. For instance, the rack heights are required to be higher than a threshold to allow objects to be inserted beneath. Their distributions become more uniform when the task is about placing the objects on top of the rack. 

%------------------------------------------------------------------------------%
\subsection{Quantitative Results in Target Tasks}
\label{sec:experiments-quantitative}

We evaluate the robot's performance in the target tasks using different methods. Our method and baselines that learn skills or generate tasks are first trained for 500K iterations without the notion of the target tasks. Then each method is trained and evaluated in the target tasks for 800k iterations. During training, we evaluate the average return of the trained model for 50 episodes every 10k iterations. 

\begin{figure*}[t!]
    \centering
    \includegraphics[width=\linewidth]{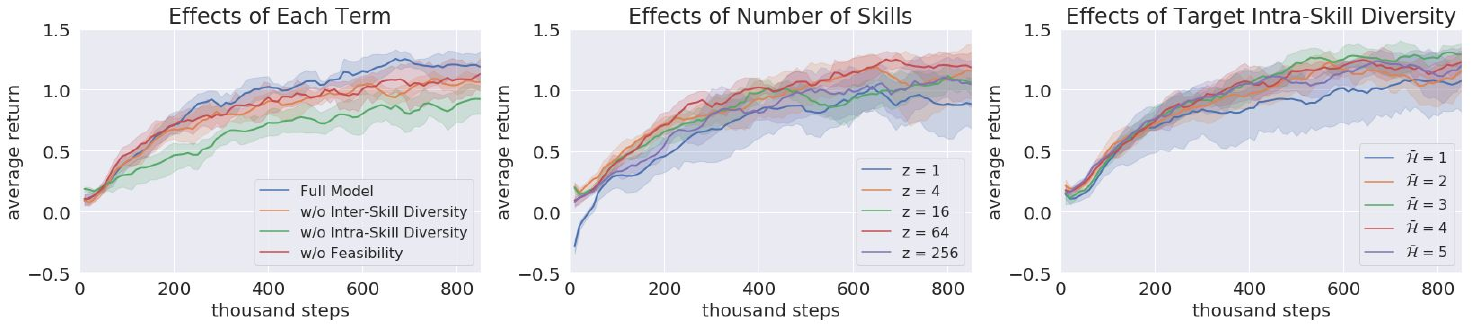}
    \vspace{-4mm}
    \caption{Ablation study. We investigate the effects of each term in the object, the number of skills, and the target intra-skill diversity. The training and evaluation of each ablation follow the same protocol as in the quantitative results and the average returns are plotted. }
    \vspace{-3mm}
    \label{fig:experiments-ablation}
\end{figure*}

\textbf{Baselines.} We compare SLIDE with multiple baseline methods which include two model-free RL algorithms trained from scratch, two skill learning algorithms, and a task generation baseline. \textbf{SAC}~\cite{Haarnoja2018SoftAO} is the state-of-the-art model-free RL baseline for continuous action spaces that are directly trained in the target task. It uses similar actor network and critic network as in our method except that the outputs are not conditioned on the skill index. \textbf{HRL}~\cite{barto2003recent} trains the same hierarchical policy described in Sec.~\ref{sec:background} except that the skill policy is jointly trained from scratch in the target tasks. \textbf{DIAYN}~\cite{Eysenbach2019DiversityIA} is the state-of-the-art method of unsupervised discovery of skills that maximizes an information-theoretic objective directly on the skill policy. Similarly, \textbf{DADS}~\cite{Sharma2020DynamicsAwareUD} aims to discover predictable behaviors by maximizing the diversity of the skills defined by a jointly learned dynamics model. We train the two skill learning baselines in randomly initialized environments sampled from the parameterized task space. We also include a task generation baseline \textbf{Uniform} which uniformly samples tasks from the parameterized task space without using a learned task generator. To have a fair comparison, we use the same network architecture for the corresponding components (\eg the high-level policy and the skill policy) and search for the optimal hyperparameters for each method. 

\textbf{Comparative analysis.} We evaluate and analyze the robot's performance in the target tasks using different methods. The evaluation across 5 runs is shown in Fig.~\ref{fig:experiments-comparison} where the curves indicate the average return and the shades indicate the standard deviations. Our method consistently outperforms the baseline methods in all target tasks in terms of the learning efficiency and the average return at convergence. The skills learned by SLIDE successfully enable the robot to solve the unseen target tasks with a high learning efficiency. The hierarchical policy effectively composes and finetunes the learned skills to interacts with different types of objects by performing various behaviors. As a comparison, the SAC baseline without using learned skills often learns quickly at the beginning of training since it has a simpler policy network to train and only focuses on a single mode of robot behavior. However, it fails to discover and utilize diverse behaviors that are required for completing the target task at a later stage. While SAC can achieve a reasonably good performance in simpler tasks that require limited modes of behavior (\eg \text{PushEnv-B} and \text{TidyEnv-B}), the performance gap between our method and SAC is much larger in more challenging tasks. The HRL baseline trained from scratch enables the robot to utilize multiple modes of behaviors. Nevertheless, HRL cannot effectively explore the environment using randomly initialized skills. The two skill learning baselines, DIAYN and DADS, can hardly to discover semantically meaningful skills in these two task domains. Instead of discovering robot behaviors such as pushing and grasping, these baselines tend to learn to move the robot gripper to different 3D locations without effective interactions with the objects on the table, which are sufficient for satisfying their diversity metrics defined on the next state. As a result, they achieve similar or even worse performance than HRL since the sub-optimal skills can lead to poor exploration in the target tasks. By randomly sampling from the task space, the Uniform baseline often struggles to provide suitable tasks to learn generalizable skills and as a result it cannot efficiently learn to solve the target tasks. 

%------------------------------------------------------------------------------%
\subsection{Ablation Study}

We run three ablation studies to investigate the importance of each component and hyperparameter to the task performance in the target tasks. Following the same training and evaluation protocols as in Sec.~\ref{sec:experiments-quantitative}, we run each ablation to learn skills in for \text{PushEnv} and then train the hierarchical policy with the obtained skills to solve the target task \text{PushEnv-A}. The resultant average return of each ablation across 5 runs is summarized in Fig.~\ref{fig:experiments-ablation}.

\textbf{Effects of each term.} To understand the importance of each term in the objective function, we train our method by removing one of the terms in the objective function (Eq.~\ref{eqn:objective_kkt}) in each ablation. As a result, the average return in the target task at convergence is reduced by 0.1 - 0.3 as shown in Fig.~\ref{fig:experiments-ablation}. Removing the intra-skill diversity leads to the largest performance regression since each generated task would collapse into a concentrated task distribution which causes the corresponding skill to overfit to a very specific scenario. Without considering the inter-skill diversity, the generated tasks associated with different skill indices tend to overlap with each other and therefore the learned skills would have poor coverage of the robot behaviors needed for solving the target task. Nevertheless, the task distribution of each skill sometimes can still become unique if they happen to be differently initialized. Since the feasibility term encourages each generated task to be solvable by the paired skill, the task distribution would still capture tasks of similar semantics at convergence. Without the feasibility term, we found that less meaningful tasks can be created by the task generator and the high-level policy is thus overwhelmed by useless skills when learning to solve the unseen target tasks.   

\textbf{Effects of number of skills.} While each task domain can inherently contain a specific set of behavior modes, we do not assume that we know how many skills can be learned beforehand. Instead, we define the number of skills to be a hyperparameter of the method. As shown in Fig.~\ref{fig:experiments-ablation}, we run our method by sweeping the number of skills from 1 to 256. On one hand, learning only a single skill leads to very limited modes of robot behaviors during the discovery of skills. As a consequence, the robot struggles to effectively explore the environment in the target task at the beginning and the performance at convergence is sub-optimal. On the other hand, setting the number too large can make the learned tasks and skills highly redundant, which confuses the model in the target task. We found the optimal number of tasks and skills is around 64 in the chosen task domain.    

\textbf{Effects of the target intra-skill diversity.} We study the effects on the target intra-skill diversity $\bar{\mathcal{H}}$ which is defined as a hyperparameter in our method. $\bar{\mathcal{H}}$ controls the balance between the intra-skill diversity and other terms in the objective. In Fig.~\ref{fig:experiments-ablation}, we report the performance by sweeping $\bar{\mathcal{H}}$ from 1 to 5. We found that using too high or too low values of $\bar{\mathcal{H}}$ can cause either the inter-skill diversity or the intra-skill diversity to outweigh each other. In our experiments, the $\bar{\mathcal{H}}$ leads to the best performance is around 3. 
\section{Conclusion and Discussion}

We present our method, Skill Learning In Diversified Environments (SLIDE), which learns generalizable skills by the automated generation of a diverse set of tasks. By maximizing the diversity of the generated tasks, our method is able to discover a variety of tasks to enable skill policies performing diverse robot behaviors to emerge. By training a hierarchical reinforcement learning algorithm that utilizes the learned skills as low-level policies, our method effectively improves the performance and the learning efficiency in unseen target tasks from two tabletop manipulation domains.   

Several aspects of the proposed method can be further investigated in future work. First, while the proposed method is designed to learn a fixed number of skills, an exciting direction would be conducting open-ended discovery of tasks and skills with a flexible number of skills. Second, we assume the parameterized reward function is predefined in the task space which suggests task goals that are potentially useful in target tasks, but future work could instead generate tasks based on intrinsically motivated reward functions. Lastly, we hope this work could encourage more endeavors in utilizing procedural content generation for robot learning and similar approaches can be proposed for a broader scope of applications such as visual navigation and humanoid robots.

\vspace{+3mm}
\textbf{Acknowledgement:} We acknowledge the support of Toyota (1186781-31-UDARO) and HAI-AWS cloud credits. 
We would like to thank Ademi Adeniji, Ajay Mandlekar, and Eric Li for their constructive feedback.

%% Use plainnat to work nicely with natbib. 

% \bibliographystyle{plainnat}
% \bibliography{references}

%% \newpage
% \clearpage
% \appendix
% \input{appendix-a-environment-details}
% \input{appendix-b-implementation-details}

\end{document}